\documentclass[letterpaper, 10 pt, conference]{ieeeconf}  

\IEEEoverridecommandlockouts                              

\usepackage{xcolor}
\usepackage{times}
\usepackage{epsfig}
\usepackage{graphicx}
\usepackage{amsmath}
\usepackage{amssymb}
\usepackage{gensymb}
\usepackage{multirow}
\usepackage{booktabs}
\usepackage{multicol}
\usepackage{caption}
\usepackage{subcaption}
\usepackage{hyperref}
\usepackage[utf8]{inputenc}

\newcommand{\ie}{\textit{i.e.}\ }
\newcommand{\eg}{\textit{e.g.}\ }
\newcommand{\etal}{\textit{et al.\ }}

\newcommand{\kilicarslan}{Kiliçarslan\ }

\usepackage{tikz}
\usepackage{textcomp}
\usepackage{hyperref}
\usepackage{lipsum}
\newcommand\copyrighttext{%
  \footnotesize \textcopyright 2023 IEEE. Personal use of this material is permitted.
  Permission from IEEE must be obtained for all other uses, in any current or future
  media, including reprinting/republishing this material for advertising or promotional
  purposes, creating new collective works, for resale or redistribution to servers or
  lists, or reuse of any copyrighted component of this work in other works.
  DOI: \href{https://doi.org/10.1109/ITSC57777.2023.10422014}{10.1109/ITSC57777.2023.10422014}}
  
\newcommand\copyrightnotice{%
\begin{tikzpicture}[remember picture,overlay]
\node[anchor=south,yshift=10pt] at (current page.south) {\fbox{\parbox{\dimexpr\textwidth-\fboxsep-\fboxrule\relax}{\copyrighttext}}};
\end{tikzpicture}%
}

\title{\LARGE \bf An object detection approach for lane change and overtake detection from motion profiles}
\author{Andrea Benericetti$^{1}$, Niccolò Bellaccini$^{1}$, Henrique Piñeiro Monteagudo$^{1,2, \dagger}$,\\ Matteo Simoncini$^{1}$, Francesco Sambo$^{1}$%
    \thanks{$^{1}$Verizon Connect Research, Florence, Italy}
	\thanks{$^{2}$DISI, University of Bologna, Italy}
	\thanks{$^{*}$Email: \tt\small andrea.benericetti@verizonconnect.com}
    \thanks{$^{\dagger} $H. P. M. acknowledges support from the SMARTHEP project, funded by the European Union’s Horizon 2020 research and innovation programme, call H2020-MSCA-ITN-2020, under Grant Agreement n. 956086}
}

\begin{document}

\maketitle
\copyrightnotice
\thispagestyle{empty}
\pagestyle{empty}

\begin{abstract}
In the application domain of fleet management and driver monitoring, it is very challenging to
obtain relevant driving events and activities from dashcam footage while minimizing the amount of information stored and analyzed. 
In this paper, we address the identification of overtake and lane change maneuvers with a novel object detection approach applied to \emph{motion profiles}, a compact representation of driving video footage into a single image.
To train and test our model we created an internal dataset of motion profile images obtained from a heterogeneous set of dashcam videos, manually labeled with overtake and lane change maneuvers by the ego-vehicle. 
In addition to a standard object-detection approach, we show how the inclusion of CoordConvolution layers further improves the model performance, in terms of mAP and F1 score, yielding state-of-the art performance when compared to other baselines from the literature.
The extremely low computational requirements of the proposed solution make it especially suitable to run in device.  
\end{abstract}

\section{INTRODUCTION}

The analysis of the footage recorded from a dashcam mounted inside a vehicle is becoming more and more important in the research literature \cite{li2022survey}.
This interest is motivated by the increase in applications that use dashcam data \cite{adamova2020dashcam},
ranging from the ones related to real-time alerting and safety \cite{ke2020edge, yu2022vehicle, pjetri2019light}, to autonomous driving \cite{wang2019monocular} and a-posteriori analysis of the recorded footage \cite{taccari2018classification, bravi2021detection, simoncini2022} (\eg used by insurance companies or in the fleet-management industry).

In recent years, with the rise of edge computing, dashcam devices have been evolving from simple recording tools to advanced driver assistance systems (ADAS), leveraging in-cabin alerts to affect driving style and pre-processing data in-device in order to save on costs related to transmission, storage and computation on the cloud \cite{ke2020edge}.
In this scenario, being able to process data in a lightweight manner is key, as it allows to save on hardware requirements. 
This can be accomplished by limiting the amount of computational operations required (\eg by using smaller ML models \cite{magistri2022}), by limiting the amount of data to process (\eg lowering the frame-rate \cite{yu2022vehicle}), or by pre-processing data and extracting relevant information to be sent to the cloud \cite{pjetri2019light}.

\begin{figure}
    \centering
    \includegraphics[width=0.48\textwidth]{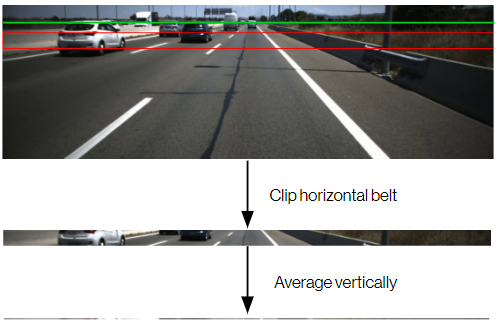}
    \caption{Frame to motion profile strip. In green, the detected horizon line, in red the considered pixel belt.}
    \label{fig:frame-to-flow}
\end{figure}

With this aim, some approaches have recently been exploring the novel and lightweight data representation known as \emph{motion profiles}.
First introduced in \cite{kilicarslan2014detecting}, this representation proposes to compress the whole video stream, possibly consisting of several seconds or minutes of video footage,  into a single 2D image.
Given a $H \times W \times T$ video, with $W$ and $H$ width and height of the frames and $T$ number of frames, a fixed strip of pixels below the horizon is considered for each frame; each strip gets then vertically averaged obtaining, in this way, a single line of pixels (see Fig.~ \ref{fig:frame-to-flow}). 
These lines are finally stacked vertically in order to generate a $W \times T$ image, where the $x$ axis represents the space domain along the lateral direction, while the $y$-axis describes how the scene unfolds over time.

\begin{figure*}
    \centering
    \includegraphics[width=0.85\textwidth]{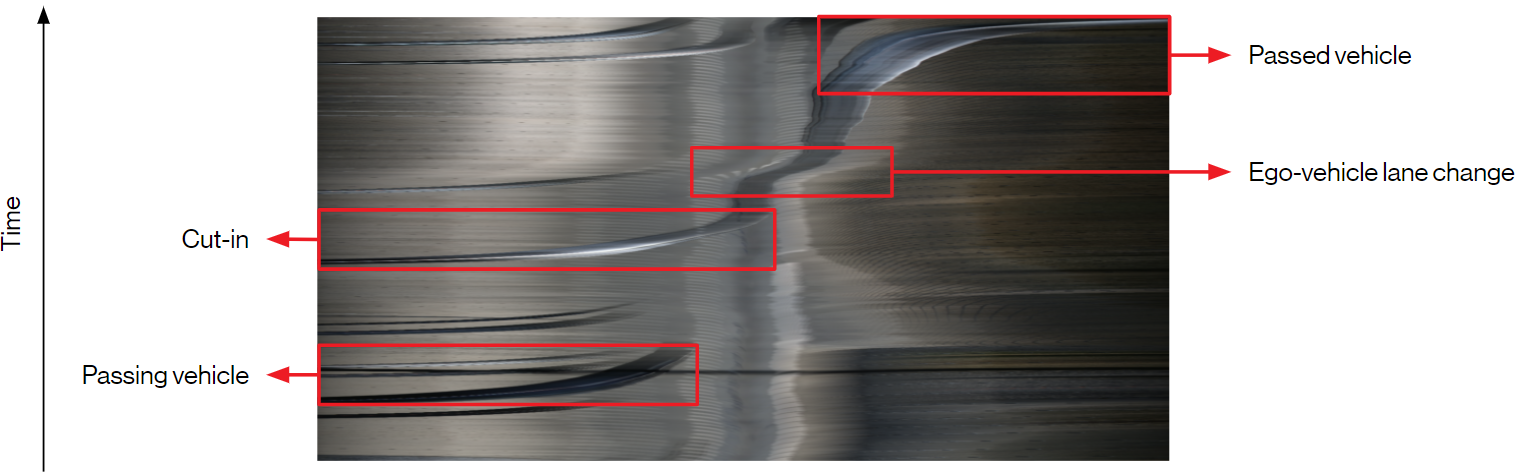}
    \caption{Example of motion profile with the resulting maneuvers annotated.}
    \label{fig:example}
\end{figure*}

Despite the heavy compression (of a factor $H$, with respect to the original video), the motion profile presents a set of unique patterns that can be used to identify events in the original video. These patterns have been studied in the literature to detect pedestrians \cite{kilicarslan2014detecting, kilicarslan2014visualizing}, oncoming vehicles \cite{kolcheck2019visual}
passing or being passed by a vehicle \cite{wang2020detecting, wang2021planning}, cut-ins \cite{wang2020detecting, wang2021planning},
presence of a leading vehicle and time-to-collision estimation \cite{wang2020detecting, kilicarslan2016bridge, kilicarslan2017direct}.
An example of such patterns is highlighted in Figure \ref{fig:example}.
The aforementioned studies generally report good performance despite the data compression, proving that the motion profile could be a valid alternative in the resource-sensitive applications that do not allow the usage of the full video.

To summarize, the motion profile is a compressed (yet, rich) representation of the road scene, 
that allows one to process a driving scene with less computational resources 
but with a trade-off on the overall performance.
In this paper, we propose to assess the suitability of the motion profile to be used in an ADAS setup based on edge-computing.
In order to do so, we selected the tasks of lane change and overtake detection (distinguishing between left and right events) and we propose to address the problem with an object detection algorithm (YOLOv3 \cite{redmon2018yolov3}), identifying the motion profile patterns of the maneuvers as objects in the spatio-temporal domain.
Moreover, we propose an extension to the classical object detection architecture by replacing the convolutional operations of the backbone of the detector with CoordConv layers \cite{liu2018intriguing}, providing the absolute position of the motion profile patterns to the network, and we show how this approach reaches a new state-of-the-art on the detection of the selected maneuvers from motion profiles.
Finally, we report the model inference time and the motion profile generation time, showing that the proposed solution allows to reach real-time detection.

The remaining of the paper is structured as follows.
In Section~\ref{sec:related-works} we present a review of the scientific literature on motion profiles and on the lane change and overtake detection tasks.
In Section~\ref{sec:methodology} we describe the proposed architecture and the enhancements to the backbone based on CoordConv layers, 
while in Section~\ref{sec:experiments} we report an extensive experimentation assessing the performance of the proposed architecture on the motion profile with respect to other baselines in the literature and highlight the strengths and limitations of the method.

\section{RELATED WORK} \label{sec:related-works}

\subsection{Maneuver Detection}
Vehicle maneuver detection is a complex task involving the analysis of temporal sequences -- usually video and sometimes other sensor data (GPS, IMU, Lidar) -- of importance for road scene understanding.

In \cite{singh2023road}, the authors propose a convolutional neural network (CNN) named \emph{3D-RetinaNet} to classify road agent actions from videos captured by a vehicle-mounted camera, building on the success of video recognition networks for human action recognition tasks like SlowFast \cite{Feichtenhofer2019slowfast}. This model receives a sequence of video frames as input and builds a feature pyramid using either a 3D or a 2D CNN as backbone; these features are then fed to two subnetworks, in order to locate other road agents and classify their actions. An additional classification head outputs ego-vehicle action classification. This architecture is fairly complex and targets a more general scenario (not only ego-vehicle maneuvers, but generic road agent actions) than ours. In addition, the authors do not report good results on ego-vehicle overtaking and moving to the sides.

Recurrent neural networks (RNNs) in combination with 2D CNNs have also been explored as a way to combine spatio-temporal information for the maneuver detection task. In \cite{peng2018driving}, features extracted from video frames through a \mbox{VGG-19} CNN are combined with handcrafted features from GPS and IMU data and fed to a LSTM-based RNN to perform ego-vehicle maneuver classification. The targeted maneuvers are left and right turns, lane changes and straight driving. While the authors report high accuracy results, this model is computationally heavy as it requires to run a deep feature extractor on each video frame, making it unsuitable for real-time execution.

Zekany \etal \cite{zekany2019classifying} explicitly target ego-vehicle maneuver classification. They propose an approach based on deep learning visual odometry to estimate ego-vehicle localization and use dynamic time warping to match the obtained trajectory against a set of reference maneuvers, namely left, right, U and K turns, stops, reverse and straight driving. This method is thus only partially applicable to our target maneuvers; moreover, the system is not designed for real-time operations and works offline, as it peeks into future frames to generate its output. Further work from the authors \cite{zekany2022finding} expands the methodology to additional maneuvers such as lane changes, decelerations, highway merges and exits by introducing individual classifiers and techniques to combine them. While this produces more general results, it also makes the system more complex, retaining the limitations for its real time operation.

While targeting tasks similar to ours, these methods rely on processing each video frame, making them computationally heavy and mostly suited for offline execution. In this work we focus on methods that use a condensed video representation and are suitable for deployment on resource constrained devices.

\subsection{Motion Profile}

The idea behind motion profiles is introduced for the first time by \kilicarslan \etal 2014a \cite{kilicarslan2014visualizing},
where the authors propose to look at a single strip of pixels for each frame at an intermediate distance between the horizon line and the vehicle hood for each frame, 
and stack them into a single image that they call road profile.
Although such representation maintains several interesting properties of the road, like pedestrian crossings and lanes, it produces several artifacts corresponding to the objects in the road scene, \eg vehicles.
For this reason, in the same paper the authors propose to look at a set of pixels under the horizon rather than a single line, to vertically average them and to form a stacked representation that they name motion profile.
They claim that in this second representation, a wide set of events relative to the motion of the entities in the scene is recognizable, 
like ego-vehicle moving, stopping, turning or changing lanes, 
as well as other vehicles passing and being passed, the presence of the leading vehicle and pedestrians; at the same time, the sharp artifacts that characterize the road profile are blurred and blend with the background.
In \kilicarslan \etal 2014b \cite{kilicarslan2014detecting} the authors address pedestrian detection and propose to consider a belt of pixels slightly below the horizon, to detect artifacts similar to chains due to the motion of the human legs; they also propose a method based on HOG to identify the nodes of such chain-like artifacts.

In \kilicarslan \etal 2016, 2017 \cite{kilicarslan2016bridge, kilicarslan2017direct} the authors use the motion profile for Time-to-collision (TTC) estimation without the need for a vehicle detection module. They propose to compute the gradient in the motion profile in the central zone of the profile, to identify the edges relative to the leading vehicle, and use convergence/divergence factors to estimate the TTC.

The first approach that tries to use motion profiles for maneuver detection is Wang \etal \cite{wang2020detecting}. 
In the paper, the authors identify a set of maneuvers and events observable in the driving scene and associate to each of them the ideal artifact that they would generate in the motion profile.
Such set includes cut-in, overtaking, being overtaken, oncoming vehicles, and left or right turns performed by the ego-vehicle.
The authors also propose a way to identify such maneuvers, by computing the gradient of the motion profile, extracting the punctual angular value of the detected traces and then computing a 1D vertical Laplacian filter into specific location of the image, \eg the center or the left or right side.
They propose to use the output of the Laplacian filter to identify the presence of a trace, and the angle value to identify its type (distinguishing for instance between passed, passing and oncoming vehicle).
Another approach for maneuver detection is presented in \kilicarslan \etal 2022 \cite{kilicarslan2022motion}.
The authors propose to consider a vehicle detected into consecutive frames at timestamp $t$ and $t + d_t$ and to compute the motion profile between the two frames, with $d_t$ reasonably small. Then, they build an image by concatenating vertically a patch with the upper part of the vehicle at frame $t$, the motion profile, and the lower part of the vehicle at frame $t + d_t$.
Finally, they use a YOLOv3 detector on the generated patches to identify both the vehicles and the traffic flow, \ie distinguishing between passed, passing and following vehicles.
Despite their approach being interesting, it does not extend to a motion profile of arbitrary length, \eg a few minutes, if not by running the inference every $d_t$ frames, which is not ideal for a lightweight setup.

\section{METHODOLOGY} \label{sec:methodology}

The aim of this paper is to leverage the motion profile concept to address the problem of maneuver detection. 
In this Section, we discuss how to build a motion profile from driving footage and highlight the advantages of this representation compared to processing the full driving video.
Moreover, we present an architecture based on YOLOv3 \cite{redmon2018yolov3} and CoordConv \cite{liu2018intriguing} that we propose to tackle the problem, further highlighting its advantages compared to a simple object detector.

\subsection{Motion profile} \label{sec:motion-profile}

The motion profile is a condensed representation of driving footage, \ie footage recorded from a camera placed on the dashboard of a vehicle with the aim of recording the road ahead.
It is computed by considering a pixel belt at a given distance below the horizon, that is vertically averaged into a single strip.
Then, each strip of each frame gets stacked to form a $T \times W$ image, with $T$ number of frames considered and $W$ width of the frames.
A schematic representation of this process is reported in Figure~\ref{fig:frame-to-flow}, while an example of motion profile is reported in Figure~\ref{fig:example}.

Such representation has many advantages. 

First, the representation greatly reduces the size of the data. 
In comparison with a video of $T$ frames of shape $W \times H$ that, thus, requires $T * W * H$ values to be elaborated, the motion profile requires just $T * W$ values, that could imply a reduction of three orders of magnitude on the size of the data in modern video resolutions (\eg $1280\times 720$).

Second, due to the condensation of the pixel belt, many small and potentially non-relevant details of the scene get blurred out, \eg small objects and shadows, while the motion of the relevant (\ie large enough) objects is preserved. 
It is worth to mention, as highlighted in \cite{kilicarslan2014detecting}, that by construction the horizontal details tend to be preserved more than the vertical ones: for instance, it is still possible to identify the width of a leading vehicle but not its height \cite{kilicarslan2016bridge}.

Third, as a result of the way it is built, everything outside of the pixel belt is ignored or heavily washed away. 
This includes everything above the horizon or too far away, \ie background information, and everything too close to the vehicle, \ie elements like the vehicle hood that generally are not relevant for road scene understanding and that the various approaches (\eg CNNs) have to learn to ignore.

Finally, the construction of the motion profile is an iterative process that perfectly suits the needs of edge computing, as the motion strip could be extracted in a lightweight manner from each frame as soon as it is acquired from the recording device and stored locally.
Performing a simple averaging operation across a 1280 pixels-long strip on modern hardware takes as little as 900 $\mu$s. Since even high-end dashcams seldom have a frame rate higher than 60 fps, extracting the motion strip is by far faster than what is imposed by real-time constraints.
In this setup, the overhead of generating the representation can be considered negligible.

A preliminary step for the motion profile generation is the identification of the horizon and the vanishing point in the scene.
Determining precisely the vanishing point coordinates $(v_x, v_y)$ in the input video is in fact crucial for the motion profile generation: its vertical position $v_y$, corresponding to the horizon, is used as a reference to identify the belt of interest as explained above. Furthermore, supposing that the dashcam is oriented towards the forward-facing direction of the vehicle, $v_x$ varies along the lateral direction of movement. Despite having a smaller impact on the motion profile, $v_x$ can be used to make some assumptions on the geometry of the scene, such as determining whether an object is to the right, to the left or in front of the ego-vehicle.

The position of the vanishing point is tightly related to how and where the dashcam is installed inside a vehicle (\ie where it is positioned on the windshield and the angle of the lens with respect to the road-facing direction), the intrinsic mechanical details of the camera (\ie field of view, potential distortions, etc.) and the characteristics of the vehicle itself (\ie its height). While in a few cases, such as controlled environments and one-vehicle settings, the vanishing point can be measured and treated as fixed, in a general setup it needs to be estimated on a per-video or per-vehicle basis: for this reason, we propose to dynamically estimate the vanishing point position with an approach similar to \cite{shuai2017regression}, through a simple 2D linear regressor, composed of a MobilenetV2 \cite{sandler2018mobilenetv2} backbone and a fully-connected layer. The model is trained with an MSE loss on a dataset of $\sim$2600 images.

Once the horizon is defined, it is needed to identify the relevant pixel belt from which to extract the motion profile. 
Intuitively, the closer the belt to the horizon, the more representative the motion profile is of objects at a far distance, and vice-versa.
Also, the higher the belt, the bigger the objects that get washed away in the final representation.
Following what is proposed in \cite{wang2020detecting}, we initially considered three belts at $\left[0,35\right]$, $\left[35,100\right]$ and $\left[100,200\right]$ pixels below the horizon line, respectively responsible for the far, medium and close distance events in the video.
Preliminary observations showed that the medium distance belt is sufficient for identifying most of the maneuvers of interest, and thus such strip is the only one considered in the proposed approach.

\subsection{Object detection + CoordConv approach} \label{sec:object detection-coordconv}

In contrast with other papers in the literature, with the notable exception of \cite{kilicarslan2022motion}, we propose to treat the identification of the patterns associated to specific maneuvers in the motion profile as an object detection task.

We believe that such an approach naturally suits the identification of the maneuvers in the motion profile, for several reasons.
First, each maneuver has a distinct and well recognizable shape: for instance the cut-in maneuvers are whiskers starting from the bottom left corner (\ie the vehicle appearing in the video) that end in the middle of the image (see Fig.~\ref{fig:example} as a reference).

\begin{figure}
    \centering
    \includegraphics[width=0.48\textwidth]{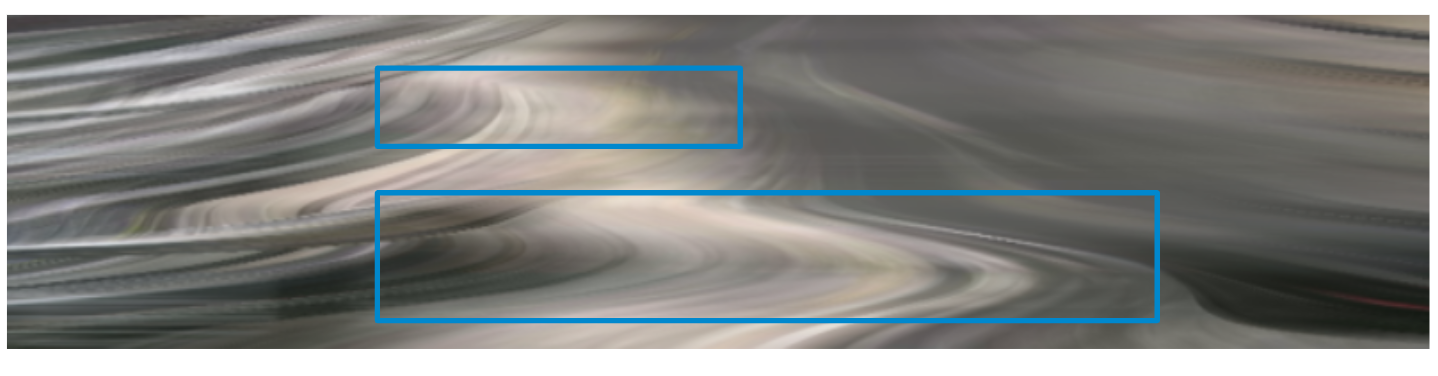}
    \caption{A motion profile with two consecutive lane changes with different shapes in time and space.}
    \label{fig:two-lane-changes}
\end{figure}

Second, despite the maneuvers having similar shape, they can be significantly different. For instance, the same cut-in event could be thinner or thicker and have a different inclination depending on the distance from the ego-vehicle, the difference of speed and on the size of the cutting-in vehicle, as shown in Figure~\ref{fig:two-lane-changes}.

What is different from a classical object detection approach is the relevance of the position: for instance, the cut-in maneuvers always end in the center of the motion profile, while the the lane changes always occur in the central part.
To fully leverage this intrinsic property, we propose to use an object detector enriched with CoordConv layers.

The CoordConv layer is first introduced in \cite{liu2018intriguing}, where the authors observe that classical convolutional architectures lack the capacity for global reasoning on the position of the object patches, as they extract features on a locally limited area and finish with a max-pooling operation. 
When processing each patch, thus, standard object detectors are unaware of its global position in the image, which is a limitation for some specific tasks, like the one we are trying to address.
Their solution is to inflate each convolutional operation with two extra channels, representing the $(x,y)$ position of each pixel in the input tensor, normalized in $\left[0,1\right]$.

Although such problem has been mitigated with the advent of Vision Transformers and Patch Position Embedding \cite{dosovitskiy2020image}, we still believe that CoordConv is a valid and simple enough solution to be applied in our scenario and we prove it with experimental results.

To tackle the detection task, we propose to replace each convolution of the backbone of a YOLOv3 network pre-trained on the COCO dataset with a CoordConv convolution, similarly to what is proposed in \cite{liu2018intriguing} for a RCNN architecture.
In order to preserve the pre-training weights, we compute an average across the channel dimension for each weight of the convolution and use such value as initialization for the newly introduced sets of weights.

\section{EXPERIMENTAL RESULTS} \label{sec:experiments}
We evaluate the performance of our proposed methodology against two baselines from \cite{wang2020detecting} and \cite{kilicarslan2022motion} by performing a series of experiments on an internal dataset.

\subsection{Dataset} \label{sec:dataset}
The dataset is composed of 416 motion profiles extracted from as many videos, following the procedure described in Section \ref{sec:methodology}. Each motion profile has a dimension of 1280 (corresponding to the width of the original video) x 480 (number of frames of the original video, which is 16 s long at 30 fps). The videos were captured from dashcams installed on the windshield of different vehicles and contain a diversity of situations (road types, urban and highway driving, etc) and conditions (day and night, weather, etc).

Four different classes are labelled in the dataset: overtakes on the left or right performed by the ego-vehicle and lane changes to the left or right done by the ego-vehicle. The 416 labelled videos contain a total of 490 events. 
For each video:
\begin{enumerate}
    \item we manually identify the start and end timestamp of each maneuver;
    \item we generate the motion profile;
    \item we convert the labels into bounding boxes in the motion profile domain: the vertical position and height of the bounding box is given by the instant the maneuver begins and ends, while the width is related to the label itself, \ie boxes related to lane changes are centered around the location of the vanishing point $v_x$ with a width of $W / 2$, while boxes related to overtakes are positioned on the side of the motion profile and extend from the edge of the image until $v_x$.
\end{enumerate}
In this way, the bounding boxes fully contain the patterns we aim to detect.

The dataset is split in train, validation and test subsets for the experiments, with a 60\% of the samples in the train set and 20\% in the validation and test sets.

\subsection{Experimental setup} \label{sec:experimental-setup}

We compare the proposed approach against two baselines from the relevant literature, namely \cite{wang2020detecting} and \cite{kilicarslan2022motion}.

While \cite{wang2020detecting} methodology aims to detect a broader number of maneuvers, they do not tackle lane change detection, thus we compare our approach only against left and right overtake detection.

In \cite{kilicarslan2022motion}, the authors propose two approaches that use object detection on motion profile: the first one generates an image composed by the top part of a frame and the bottom part of a subsequent frame, separated by the motion profile between the two, while the second one is the same with the full-frame information blacked out. 

We compare our method to the latter approach, as we argue that the first one is not suitable for an edge-computing setup:
first, it would require either to run inference every fixed amount of time (\ie when the full-frame is stored) or to store all the full-frames, falling back into the video processing problem and erasing the compact representation benefit of using motion profiles;
second, if the distance between the two frames is massive, \eg a minute, the information of the two full-frames is not relevant.

Both the proposed approach and the baseline based on \cite{kilicarslan2022motion} are trained using an ADAM optimizer with a learning rate of $10^{-3}$ and a weight decay of $10^{-3}$ for a maximum of 20 epochs.

\subsection{Results} \label{sec:results}
We summarize the results of our experiments in terms of Average Precision for detecting lane changes and overtakes in Table \ref{tab:results_ap} for our method and the baseline based on \cite{kilicarslan2022motion}, which approach the problem as an object detection task. Our method provides the highest mean Average Precision -- a 35.4\% -- performing better than the baselines by a significant margin. It performs at around 40\% Average Precision for the detection of all the maneuvers except overtakes on the left. 
\begin{table}[]
    \centering
    \begin{tabular}{cccccc}
    \toprule
         & LR & LL & OR & OL & mAP  \\
    \midrule
         Ours & 0.391 & \textbf{0.400} & \textbf{0.402} & \textbf{0.225} & \textbf{0.354} \\
         \kilicarslan et al. \cite{kilicarslan2022motion} & \textbf{0.511} & 0.398 & 0.254 & 0.075 & 0.310\\
    \bottomrule     
    \end{tabular}
    \caption{Comparison of the Average Precision of our proposed method and the object detection based baseline on the targeted maneuvers: Lane changes to the Right (LR), Lane changes to the Left (LL), Overtakes on the Right (OR), Overtakes on the Left (OL)}
    \label{tab:results_ap}
\end{table}

In Table \ref{tab:result_pr} we report the precision, recall and F1 scores of our method and the baselines on the test set. At this operating setting our method shows a high precision ($>$ 90\%) on lane changes to both sides, while maintaining a reasonable recall ($>$ 60\%). Our method performs worse on overtakes, with the precision dropping to about 50\% with a lower recall. Our method outperforms both baselines in terms of F1 score, with the method by \kilicarslan \etal providing similar results in terms of lane change detection and both baselines being inferior in overtake detection performance, especially in the case of overtakes to the left of the ego-vehicle.

An analysis of the failure cases in overtake detection reveals that the precision of the method decreases due to false positive detections produced by situations which the model cannot disambiguate. One of such situations is passing parked cars, which causes the model to sometimes predict a non-existing overtake. Another circumstance that produces false positives is the presence of vehicle queues in the opposite direction. 
On the other hand, we identify the sources that produce a decrease on recall: missing detections, \ie false negatives, when the ego-vehicle speed is much faster than an overtaken car and when the ego-vehicle overtakes cars in queues, which are sometimes mistaken with parked cars and not detected as an overtake. 

\begin{table}[]
    \centering
    \begin{tabular}{cccccc}
    \toprule
      Method & Metric & LR & LL & OR & OL \\
    \midrule
     \multirow{2}{*}{Ours} & Precision & 0.95 & \textbf{0.92} & 0.50 & \textbf{0.57} \\
     &Recall & \textbf{0.64} & \textbf{0.77} & \textbf{0.50} & \textbf{0.28} \\
     &F1 score & \textbf{0.76} & \textbf{0.84} & \textbf{0.50} & \textbf{0.38} \\
     \midrule
     \multirow{2}{*}{\kilicarslan et al. \cite{kilicarslan2022motion}} & Precision & \textbf{1.00} & 0.82 & \textbf{0.57} & 0.35 \\
     & Recall & 0.61 & 0.70 & 0.42 & 0.12 \\
     &F1 score & \textbf{0.76} & 0.76 & 0.48 & 0.18 \\
     \midrule
     \multirow{2}{*}{Wang et al. \cite{wang2020detecting}} & Precision & - & - & 0.10 & 0.14 \\
     & Recall & - & - & 0.27 & 0.22 \\
     & F1 score & - & - & 0.15 & 0.17 \\
    \bottomrule     
    \end{tabular}
    \caption{Performance on the test set. Fixed intersection over union threshold of 0.3 and fixed confidence threshold of 0.2 for the two object detection based methods}
    \label{tab:result_pr}
\end{table}

The inference time of the proposed approach is 70 ms on GPU (computed on a NVidia V100) and 400 ms on CPU (computed on an Intel Xeon @ 2.1 GHz) by considering a motion profile of 16 seconds.
These results, along with the fact that it is sufficient to run the maneuver detection on motion profile at low frequency (\eg every few seconds), make the proposed approach suitable for an edge-computing scenario.

\section{CONCLUSIONS AND FUTURE WORK} \label{sec:conclusions}

In this paper, we propose a lightweight solution to detect left/right ego-vehicle overtakes and lane changes on the road from dashboard camera footage. The solution leverages a compact representation of the video footage as a single image, called motion profile. On such image, we run an object detector to detect the graphic signatures of the four maneuvers. The object detector is made aware of the object position within the image through a CoordConv layer. Our method shows state-of-the-art detection results in all the maneuvers we considered. Our model performs significantly better on lane change than on overtake, probably due to the intrinsic difficulty of distinguishing the latter from parked cars in a real-world scenario. 

As future improvements, we plan to measure the impact of more recent and efficient detectors on our pipeline, as well as expand our methodology to a broader set of events detectable within the motion profile, like cut-ins or presence of a leading vehicle. 
Moreover, we want to tackle the intrinsic ambiguity of the parked vehicles misdetected as overtakes, to further improve the overall performance.

\addtolength{\textheight}{-12cm}   

\bibliographystyle{IEEEtran}
\bibliography{IEEEabrv,biblio}

\end{document}